\documentclass[journal]{IEEEtran}
\usepackage[utf8]{inputenc}
\usepackage{amssymb}
\usepackage{graphicx}
\usepackage{booktabs}
\usepackage{multirow}
\usepackage{cite}
\usepackage{subfigure}
\usepackage{enumerate}
\usepackage{threeparttable}
\usepackage[switch]{lineno} 
\usepackage{lipsum} 
\usepackage{colortbl}
\usepackage{hhline}

\newcommand{\cb}[1]{{\color{black} #1}}
\newcommand{\cc}[1]{{\color{blue} #1}}

\ifCLASSOPTIONcompsoc
  \usepackage[caption=false,font=normalsize,labelfont=sf,textfont=sf]{subfig}
\else
  \usepackage[caption=false,font=footnotesize]{subfig}
\fi
\begin{document}

\title{Multitask Deep Learning with Spectral Knowledge for Hyperspectral Image Classification}

\author{Shengjie Liu, and Qian Shi
\thanks{This work was supported by the National Natural Science Foundation of China under Grant 61601522. \emph{(Corresponding author: Qian Shi.)}} 
\thanks{S. Liu is with Department of Physics, University of Hong Kong, Hong Kong SAR, China, and was with OneSpace Technology Co., Ltd., Chongqing 400020, China and Guangdong Provincial Key Laboratory of Urbanization and Geo-simulation, Sun Yat-sen University, Guangzhou 510275, China  (e-mail: liushengjie0756@gmail.com).}
\thanks{Q. Shi is with Guangdong Provincial Key Laboratory of Urbanization and Geo-simulation, School of Geography and Planning, Sun Yat-sen University, Guangzhou 510275, China (e-mail: shixi5@mail.sysu.edu.cn).}
}

\markboth{Accepted by IEEE GRSL}%
{Accepted by IEEE GRSL}

\maketitle

\begin{abstract}
\cc{This is the preprint version. To read the final version please go to \textit{IEEE Geoscience and Remote Sensing Letters} on IEEE Xplore.}
In this letter, we propose a multitask deep learning method for classification of multiple hyperspectral data in a single training. Deep learning models have achieved promising results on hyperspectral image classification, but their performance highly rely on sufficient labeled samples, which are scarce on hyperspectral images. However, samples from multiple data sets might be sufficient to train one deep learning model, thereby improving its performance. To do so, we trained an identical feature extractor for all data, and the extracted features were fed into corresponding Softmax classifiers. Spectral knowledge was introduced to ensure that the shared features were similar across domains. Four hyperspectral data sets were used in the experiments. We achieved higher classification accuracies on three data sets (Pavia University, Pavia Center, and Indian Pines) and competitive results on the Salinas Valley data compared with the baseline. Spectral knowledge was useful to prevent the deep network from overfitting when the data shared similar spectral response. The proposed method \cb{tested on two deep CNNs}
successfully \cb{shows its ability to} utilize samples from multiple data sets \cb{and enhance networks' performance.}
\end{abstract}


\IEEEpeerreviewmaketitle

\section{Introduction}
Remote sensing image classification, also known as semantic segmentation in computer vision, provides land use and land cover information that is essential for environment and urban management. 
Thus, many supervised algorithms have been proposed to accurately discriminate land cover classes in remote sensing images, including support vector machines, random forests, and neural networks \cite{fauvel2012advances, mountrakis2011support, yoshida1994neural, melgani2004classification, ham2005investigation, Qi2012,  Chen2014}. 
Recently, deep learning, particularly deep convolutional neural networks (CNNs), has achieved considerable progress in remote sensing image classification, including hyperspectral image classification, and has achieved state-of-the-art results \cite{chan2015pcanet, zhao2016spectral, Lee2017, zhong2018spectral, haut2018active, huang2018urban}.

Hyperspectral data are rich in spectral information and thus have a high potential for land use and land cover mapping, \cb{both in soft classification (unmixing) and hard classification \cite{hong2018augmented, camps2014advances}.}
\cb{In recent years, the recurrent neural networks are regarded as effective in learning discriminative features from spectral signals while their integration with CNNs is effective in both spectral and spatial domains \cite{hang2019cascaded}}.
For semantic segmentation of hyperspectral images, a deep CNN will have more parameters if the input data have more spectral bands. 
For example, a deep Contextual CNN \cb{(CCNN hereafter)} for hyperspectral image classification was designed in \cite{Lee2017}; the number of parameters varied from 613,000 for 103 bands to 1,876,000 for 224 bands. Although the performance of the CCNN was promising, sufficient training samples were needed to prevent the network with huge amounts of parameters from overfitting.
However, collecting samples for hyperspectral data is time consuming and labor intensive, resulting limited labeled samples available on hyperspetral images.

Several methods were proposed to tackle the dilemma between big networks and small samples. 
Some studies lightened the network by utilizing the  1$\times$1 convolutional layers \cite{paoletti2018new, liu2018wcrn}. \cite{pan2018mugnet} reduced the parameters by designing a simplified four layers network without hyperparameters, which applied an ensemble manner to utilize spatial-spectral information from hyperspectral data  using 20 labeled samples per class. 
However, a deeper network with more parameters is expected to have a better performance. 
Thus, how to maintain the deepness and complexity of deep learning model while avoiding overfitting becomes the topic of interest.

One may consider using transfer learning to leverage samples from external sources. For instance, \cite{shi2015domain} attempted to find new representations for each class from the source domain to the target domain by multiple linear transformations with low-rank reconstruction. 
\cite{kim2010adaptive} first trained a kernel machine with labeled data, which was then adapted to new data with manifold regularization. \cite{persello2015kernel} proposed a kernel-based measure of data shift to select domain-invariant discriminant features for hyperspectral images. 
The aforementioned domain adaptation techniques successfully utilized samples from other source to increase models' generalization ability in the target domain, but they required the source and target data share an identical classification system, which limits the use of these methods.
\cite{yang2017learning} applied the fine-tuning technique to extract spectral-spatial features from a CNN with two branches. In a fine-tuning fashion, they were able to use samples from data sets with diverse classification systems and from different sensors. In an unsupervised manner, \cite{Kemker2017} used self-taught learning on large amounts of unlabeled samples to learn CNN models (autoencoders) that were generalized enough to extract features for supervised classification with limited labeled samples. Although labeled samples are scarce on single hyperspectral image, a lot of hyperspectral data are available in the last two decades thanks to the development of hyperspectral imaging.
Thus, if we can use labeled samples from multiple data sets in a supervised manner, we might alleviate the overfitting problem of deep CNN models.

In this letter, we propose a multitask deep learning method to leverage limited labeled samples from multiple data sets. Multitask learning is a type of transfer learning, but unlike fine tuning, multitask learning does not require the source data have a large quantity of training samples. Instead, we use samples from multiple data sets to cotrain one network, thereby improving its generalization. 
The intuition of multitask learning is that two or multiple tasks share the similar features. For hyperspectral image classification, the spectral information has its physical meaning (radiance or reflectance) and is independent of the data set used. Therefore, utilizing samples among data sets with varying classification systems and even different sensors is possible. Besides, obtaining one deep CNN model for classification of two hyperspectral data sets in a single training can save half the training time in theory \cite{Zbakh2019}, while with more data sets cotrained the more training time can be saved.
The major contributions of this letter are summarized as follows.

\begin{itemize}
\item We propose a multitask deep learning method for hyperspectral image classification that can leverage training samples from multiple data sets \cb{and achieve better machine generalization.} \cb{Also, obtaining a multitask network  in a single training saves the training time for multiple data sets.}
\item \cb{We confirm that spectral knowledge is crucial for multitask learning. Similar spectral signals provide a more robust multitask learning environment.}
\end{itemize}

\section{Methodology}
\label{sec:Methodology}

\subsection{Deep CNNs}

\begin{figure*}[!t]  
\centering      
\includegraphics[width=16.5cm]{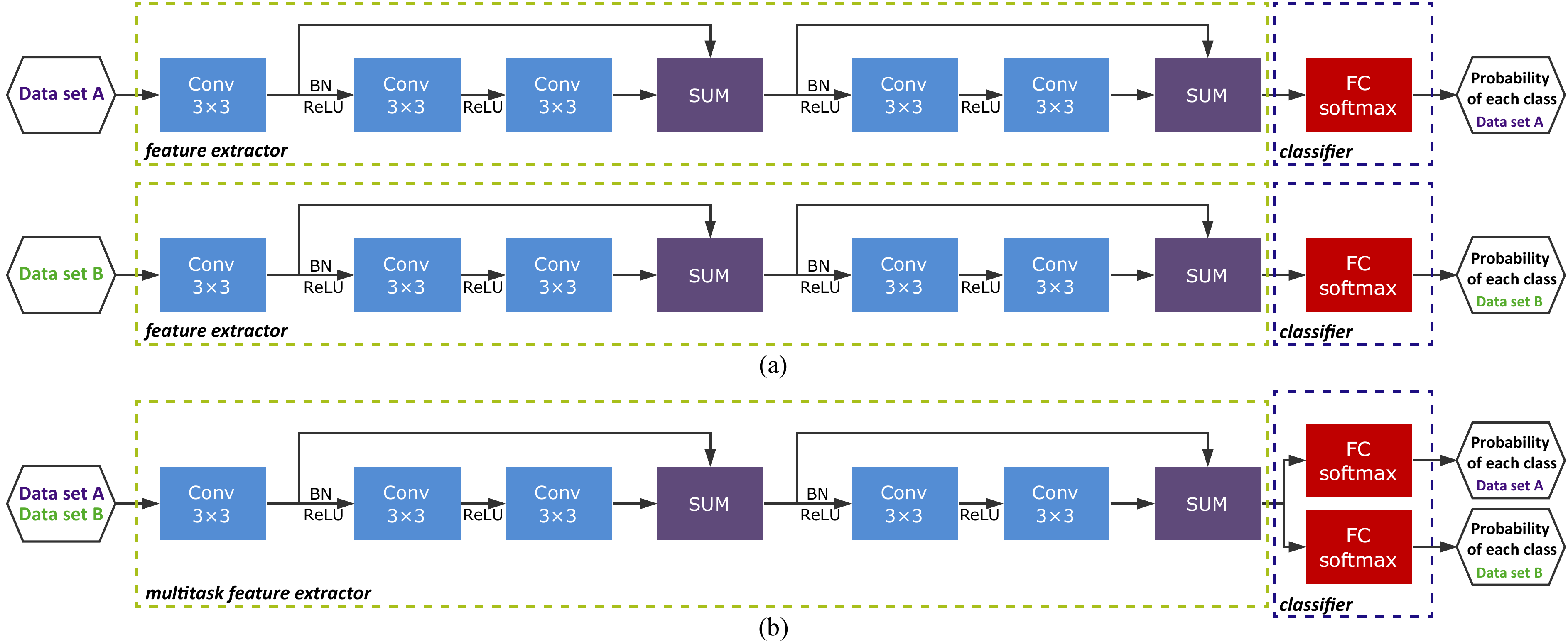}
\caption{The architecture of the network (HResNet) used in this study. \cb{(a)} \cb{HResNet, single-task version.} \cb{(b)} \cb{HResNet, multitask version.}}
\label{fig:network}
\end{figure*}

\cb{To illustrate the idea of multitask learning, we first review what a single-task network is doing.}
Deep CNNs have achieved great success in hyperspetral image classification mainly because of two reasons. One is that it can generate spatial features from the input image by convolutional layers; the other is that the generated spatial features and the spectral features of the input image (the spatial-spectral features) are transformed nonlinearly into a high dimensional feature vector, from which the classifier can easily distinguish different classes. Therefore, the structure of a deep CNN can be seen as the combination of a feature extractor {$\phi(\cdot)$} and a classifier {$f(\cdot)$} (Fig. \ref{fig:network}a). The shallow layers are treated as the feature extractor, while the last fully connected layer with the softmax function is treated as the classifier. For a input image sample $x$, a tensor with the shape of $H \times W \times C$ in our study ($H, W$, and  $C$ stand for height, width, and channels of the input image, respectively), the feature extractor $\phi(\cdot)$ extracts, fuses, and transforms the sample $x$ to a feature vector $x_{\phi}$ that the classifier {$f(\cdot)$} can better differentiate:
\begin{equation}
x_{\phi}=\phi(x).
\end{equation}

The classifier $f(\cdot)$ takes the output vector $x_{\phi}$ from the feature extractor $\phi(\cdot)$ as its input. In a CNN, the fully connected layer with the Softmax activation function serves as the classifier $f(\cdot)$ and gives the probability $P(y=j|x_{\phi})$ of the $j$-th category:
\begin{equation}
P \left(y=j|x_{\phi} \right) = \frac{\exp \left(x_{\phi}^T w_{j}+b_j \right)}{\sum_{k=1}^K{\exp \left(x_{\phi}^T w_k + b_k \right)}},
\end{equation}
where $w_j$ is the weight vector of the $j$-th neuron in the fully connected layer, $b_j$ is a bias element corresponding to the $j$-th neural, and $K$ is the number of category.

\subsection{Multitask Learning}
\cb{For multitask learning,}
let $X_p$ be the first target instance space, where each instance $x_p\in X_p$ and $p$ represents the target space. Given a labeled training data set with index $i$ consisting of a few pairs $(x_p^i,c_p^i)$, where $x_p^i \in X_p$ and $c_p^i \in C_p=\{1,...,|c_p|\}$ is the class label of $x_p^i$. The classification task for the data is to learn a deep CNN with a feature extractor ${\phi}_p(\cdot)$ and a predictive function $f_p(\cdot)$ that can predict the corresponding label $c_p$ of a new instance $x_p$. The classification task for the second target data $X_q$ is the same {\textendash} to learn a deep CNN with a feature extractor $\phi_q(\cdot)$ and a predictive function $f_q (\cdot)$. 

When samples are limited, due to the large amount of parameters in the deep CNN $({\phi}_p(\cdot),f_p(\cdot))$, especially in the feature extractor ${\phi}_p(\cdot)$, the model is easily overfitting, degrading its performance. For example, the number of parameters of the feature extractor \cb{of CCNN} is 628,736, whereas the number of parameters of the classifier is 129$\times |c_p|$, where $|c_p|$ is the number of classes.
We can tackle this issue by using one feature extractor for multiple data sets, i.e., using samples from different domain to train a deep learning model. 
The theory behind multitask learning for classification of hyperspectral data is that, 
Earth observation data should share similar spectral-spatial features, because they represent the radiance or reflectance of ground targets, which is independent of the data set used.
Thus, in \cb{multitask setting}, one feature extractor ${\phi}(\cdot)$ is enough for both or even multiple data sets \cb{(Fig. \ref{fig:network}b)}, lightening the network:
\begin{equation}
{\phi}(\cdot)={\phi}_p(\cdot)={\phi}_q(\cdot).
\end{equation}
Therefore, we can train a multitask deep CNN ($\phi(\cdot),f_p(\cdot),f_q(\cdot)$) for both data sets. Now that the training samples are enlarged, the performance of the deep model should be enhanced.

\cb{
The network (Fig. \ref{fig:network}a) we proposed in the study is referred as HResNet (Hyperspectral-ResNet) hereafter. In Fig. \ref{fig:network}b, we present its multitask version (two tasks), which has another fully connected layer with the Softmax function for the classification of a second data set. 
The performance of HResNet is presented in Tables \ref{tab:oa_pu}, \ref{tab:oa_pc}, \ref{tab:oa_in}, \ref{tab:oa_sa} for each data set, respectively. It outperformed SVM in most cases and was competitive compared to the CCNN.
Note that the proposed multitask deep learning is a method that can be integrated with any network with a large amount of parameters and the backbone network here should play a very limited role. 
To further confirm the generalization of the proposed method, we also integrate multitask learning with another powerful network, the aforementioned CCNN \cite{Lee2017}.}

\section{Experiments}
\subsection{Hyperspectral Data Sets}
Four hyperspectral data sets were used in the experiments. The first one is the Indian Pines (IN) data set with a size of 145$\times$145$\times$200 captured by  Airborne Visible/Infrared
Imaging Spectrometer (AVIRIS) sensor over an agricultural area. There are 16 classes in this data. 
The second data set, Salinas Valley (SA),  is also captured by AVIRIS sensor. It consists of a 512$\times$217 agricultural area with 204 spectral bands and 16 classes. 
The third data set is the Pavia University (PU) data set with a size of 610$\times$340$\times$103 and 9 classes. It was acquired by the Reflective Optics System Imaging Spectrometer (ROSIS) sensor over an urban area.
The last one is the Pavia Center (PC) data set, captured by the ROSIS as well. It consists of 1096$\times$715 pixels with 102 spectral bands and comprises 9 classes. 

\subsection{Experimental Setup}
We implemented the \cb{networks} using Keras with TensorFlow backend. 
The experiments were conducted on a machine equipped with a 3.0 GHz Intel Core i5-8500 CPU, 32G RAM, and an NVIDIA GeForce GTX1060 6G GPU. When training \cb{in multitask setting}, we used the AdaDelta optimizer with a batch size of 20. The learning rate was set as 1.0 for multitask learning in the first 100 epochs. Then, the learning rate was set as 0.1 to adjust to each task for 30 epochs. The final classification was voted from five predictions, which were obtained from the end of training in a step of two epochs. 
With the spectral knowledge, the first convolutional layer is weight sharing; without the spectral knowledge, each data set has its own weights of the first convolutional layer.
\cb{We also inverted spectral bands of the second data set in multitask setting to see the effectiveness of spectral knowledge.}
All the data were augmented 4 times by mirroring each sample horizontally, vertically, and diagonally.

\cb{As multitask learning can be integrated with any network with a large amount of parameters, we integrate multitask learning with another state-of-the-art CNN architecture, the CCNN \cite{Lee2017}, to illustrate the generalization of the proposed method. The CCNN is a network with up to 1,876,000 paramters (for a 224-band hyperspectral image) and was proved to be very powerful in hyperspectral image classification when samples were sufficient. Here, we'd like to test its ability under the context of very limited training samples and see if multitask learning can improve its classification. }

\begin{table*}[!t]
  \centering
  \caption{Overall accuracy along with standard deviation of Pavia University.}
  \scalebox{0.78}[0.78]{
    \begin{tabular}{c|cc|cccc|cccc|cc} \hline
    \toprule
    NoS/class  & \multicolumn{2}{c|}{\cb{Single-task}}  & \multicolumn{4}{c|}{MTL w/ spectral knowledge, \cb{HResNet}} & \multicolumn{4}{c|}{MTL w/o spectral knowledge, \cb{HResNet}} & \multicolumn{2}{c}{\cb{CCNN}}\\
  per domain &  \cb{SVM} & \cb{HResNet}    & PC    & SA    & IN    & ALL   & PC    & SA & IN & \cb{Inverse-PC}  &  \cb{Single-task} & \cb{MTL-PC}\\
    \midrule
    5   & \cb{62.17$\pm$9.08} &  62.80$\pm$7.03 & 66.11$\pm$6.30 & 64.78$\pm$3.00 & 64.84$\pm$7.04 & 65.66$\pm$6.85 & 64.25$\pm$5.31 & 64.54$\pm$5.80 & 64.45$\pm$7.18 & \cb{64.26$\pm$6.47}  & \cb{64.77$\pm$6.55} & \cb{66.23$\pm$3.73} \\
    10  & \cb{68.32$\pm$6.53} & 72.46$\pm$4.07 &75.11$\pm$3.22 & 75.07$\pm$4.37 & 74.78$\pm$5.49 & 75.88$\pm$4.76 & 74.83$\pm$4.56 & 74.39$\pm$3.87 & 74.58$\pm$3.57 & \cb{74.14$\pm$3.48}  & \cb{75.63$\pm$4.50} &  \cb{76.53$\pm$3.47}   \\
    15  & \cb{73.60$\pm$3.43} & 80.75$\pm$2.72  &82.71$\pm$3.42 & 83.92$\pm$2.43 & 82.48$\pm$3.30 & 83.09$\pm$2.71 & 83.55$\pm$2.92 & 83.53$\pm$3.56 & 82.08$\pm$3.61 & \cb{81.39$\pm$3.34} & \cb{83.05$\pm$2.15} & \cb{83.13$\pm$3.15}   \\
    20  & \cb{77.57$\pm$3.88} & 84.67$\pm$1.87 & 86.72$\pm$1.99 & 86.93$\pm$1.44 & 86.18$\pm$1.62 & 85.71$\pm$1.38 & 86.07$\pm$2.01 & 87.15$\pm$1.65 & 86.42$\pm$1.41 & \cb{84.70$\pm$3.63}  & \cb{86.58$\pm$2.31} &  \cb{86.87$\pm$1.36} \\
    25  & \cb{78.51$\pm$4.14} & 87.39$\pm$1.51 &88.88$\pm$1.43 & 89.53$\pm$1.16 & 89.09$\pm$1.82 & 88.21$\pm$0.93 & 89.20$\pm$1.07 & 89.30$\pm$1.46 & 88.61$\pm$1.09 & \cb{87.51$\pm$2.02}  &   \cb{88.52$\pm$1.80} &  \cb{88.29$\pm$1.35} \\
    30  & \cb{81.83$\pm$2.49} & 87.49$\pm$1.31 &89.82$\pm$1.41 & 90.44$\pm$1.41 & 90.68$\pm$1.16 & 88.59$\pm$1.21 & 90.28$\pm$1.27 & 90.23$\pm$1.62 & 90.21$\pm$1.20 & \cb{88.92$\pm$1.12}  & \cb{90.45$\pm$1.36} & \cb{90.51$\pm$1.50}  \\
    \bottomrule \hline
    \end{tabular}}%
  \label{tab:oa_pu}%
\end{table*}%

\begin{table*}[!t]
  \centering
  \caption{Overall accuracy along with standard deviation of Pavia Center.}
  \scalebox{0.78}[0.78]{
    \begin{tabular}{c|cc|cccc|cccc|cc} \hline
    \toprule
    NoS/class  & \multicolumn{2}{c|}{\cb{Single-task}} & \multicolumn{4}{c|}{MTL w/ spectral knowledge, \cb{HResNet}} & \multicolumn{4}{c|}{MTL w/o spectral knowledge, \cb{HResNet}} & \multicolumn{2}{c}{\cb{CCNN}}\\
    per domain & \cb{SVM} & \cb{HResNet} & PU    & SA    & IN    & ALL   & PU    & SA & IN & \cb{Inverse-SA}  &  \cb{Single-task} & \cb{MTL-SA} \\
    \midrule
    5  & \cb{90.45$\pm$3.86}  & 95.20$\pm$1.66 &   94.39$\pm$1.86  & 96.30$\pm$1.79 & \multicolumn{1}{l}{95.59$\pm$0.91} & 93.01$\pm$2.73 & 95.52$\pm$1.70 & 96.01$\pm$1.98 & 95.05$\pm$2.13  & \cb{94.17$\pm$1.63} &     \cb{92.75$\pm$2.35}  &   \cb{93.63$\pm$2.17} \\
    10  & \cb{93.84$\pm$1.06} & 96.59$\pm$1.01 &95.95$\pm$1.40 & 97.26$\pm$0.61 & 96.83$\pm$0.50 & 95.92$\pm$0.75 & 96.73$\pm$0.79 & 97.02$\pm$0.77 & 96.56$\pm$1.11  & \cb{95.80$\pm$1.43}   & \cb{94.92$\pm$1.12} & \cb{95.80$\pm$0.72} \\
    15  & \cb{94.94$\pm$0.79} & 97.46$\pm$0.23 & 97.26$\pm$0.29 & 97.64$\pm$0.23 & 97.40$\pm$0.22 & 96.95$\pm$0.46 & 97.45$\pm$0.39 & 97.54$\pm$0.32 & 97.39$\pm$0.23 & \cb{96.87$\pm$0.45}  &  \cb{95.96$\pm$0.87}  & \cb{96.58$\pm$0.41} \\
    20  & \cb{95.32$\pm$0.53} & 97.65$\pm$0.37 & 97.50$\pm$0.34 & 97.73$\pm$0.26 & 97.66$\pm$0.23 & 97.17$\pm$0.32 & 97.70$\pm$0.35 & 97.75$\pm$0.27 & 97.70$\pm$0.24 & \cb{97.24$\pm$0.40}  & \cb{96.86$\pm$0.74} & \cb{97.14$\pm$0.60} \\
    25  & \cb{95.84$\pm$0.48} & 97.82$\pm$0.31 &97.72$\pm$0.29 & 97.90$\pm$0.23 & 97.81$\pm$0.22 & 97.53$\pm$0.21 & 97.91$\pm$3.55 & 97.88$\pm$0.29 & 97.88$\pm$0.24 & \cb{97.48$\pm$0.26}  & \cb{96.92$\pm$0.78}  & \cb{97.15$\pm$0.67}  \\
    30  & \cb{95.66$\pm$0.70} & 98.02$\pm$0.20 &97.92$\pm$0.23 & 98.18$\pm$0.22 & 97.95$\pm$0.20 & 97.71$\pm$0.28 & 98.14$\pm$0.22 & 98.20$\pm$0.25 & 97.97$\pm$0.23 & \cb{97.84$\pm$0.16} & \cb{97.10$\pm$0.71}  & \cb{97.32$\pm$0.67}  \\
    \bottomrule \hline
    \end{tabular}}%
  \label{tab:oa_pc}%
\end{table*}%

\subsection{Experiments on the ROSIS Data Sets}
In the first group of experiments, we conducted multitask learning on the ROSIS data sets. 
Although the ROSIS data sets are from the same sensor, they were preprocessed individually, resulting a scene (Pavia Center) with 102 bands and the other (Pavia University) with 103 bands. To align the spectral bands, the last band of Pavia Center was repeated, so the number of input channels was 103. We used bands from 11 to 113 from the AVIRIS data sets when they were involved to train the ROSIS data sets. The overall accuricies (OAs) of Pavia University are presented in Table \ref{tab:oa_pu}. 
Compared with \cb{single-task learning}, multitask learning generated results with higher OAs in all cases. 
For example, with only 5 samples per class, the OA of multitask learning is 66.11\%, 3.31\% higher than the baseline.
Multitask learning with Pavia Center and multitask learning with another data set with a different sensor did not significantly affect its performance, indicating the robustness of multitask learning. 
When spectral knowledge was involved, the multitask model generalized better with fewer training samples. 
\cb{When we inverted the spectral bands (Inverse-PC), the performance of multitask learning was degraded, though it still improved the classification accuracy. It illustrates the importance of spectral knowledge in multitask learning.
We also conducted an experiment using CNN and the result shows that in multitask learning, CCNN achieved better machine generalization in terms of accuracy and stability (standard deviation).
}

The results of Pavia Center are presented in Table \ref{tab:oa_pc}. This data set is easy to classify, and we achieved over 95\% OA with only 5 samples per class. A higher OA was obtained by multitask learning with the Salinas Valley data. Its performance was inline with the baseline when cotraining with other data.  

\begin{figure*}[htbp]
    \centering
    \subfigure[\scriptsize{\cb{Single-task}}]{\includegraphics[width=0.118\textwidth]{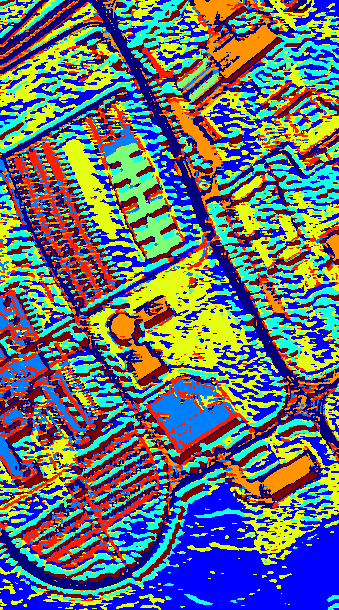}} 
    \subfigure[\scriptsize{\textbf{w/} PC}]{\includegraphics[width=0.118\textwidth]{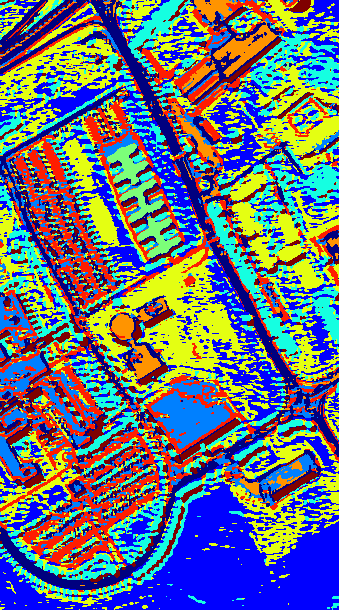}} 
    \subfigure[\scriptsize{\textbf{w/} IN}]{\includegraphics[width=0.118\textwidth]{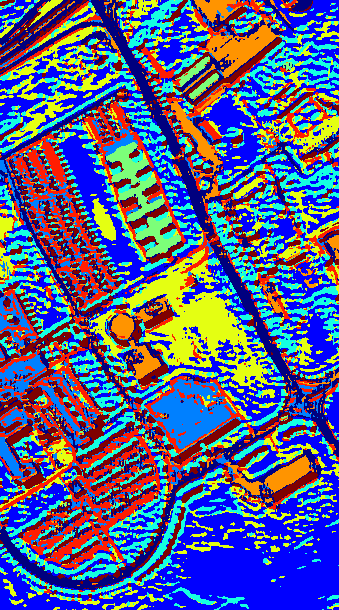}}
    \subfigure[\scriptsize{\textbf{w/} SA}]{\includegraphics[width=0.118\textwidth]{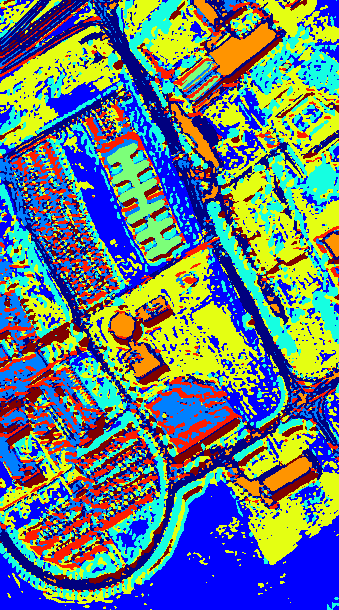}}
    \subfigure[\scriptsize{\textbf{w/} ALL}]{\includegraphics[width=0.118\textwidth]{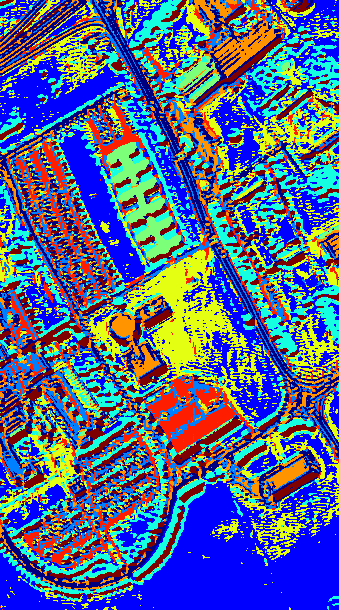}}
    \subfigure[\scriptsize{\textbf{w/o} PC}]{\includegraphics[width=0.118\textwidth]{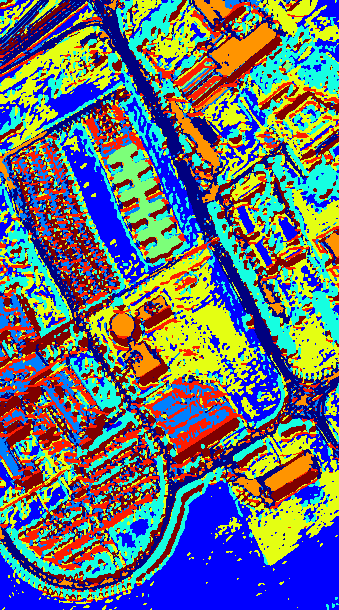}}
    \subfigure[\scriptsize{Ground Truth}]{\includegraphics[width=0.118\textwidth]{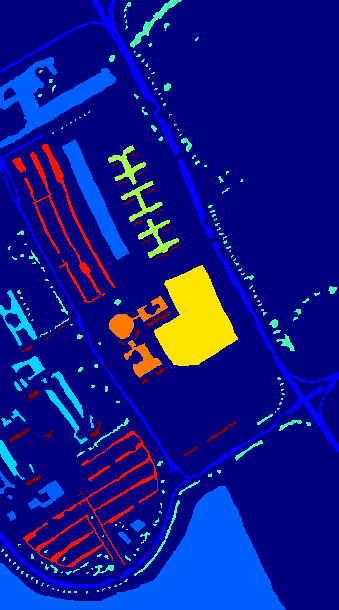}}
    \subfigure[\scriptsize{False Color Map}]{\includegraphics[width=0.118\textwidth]{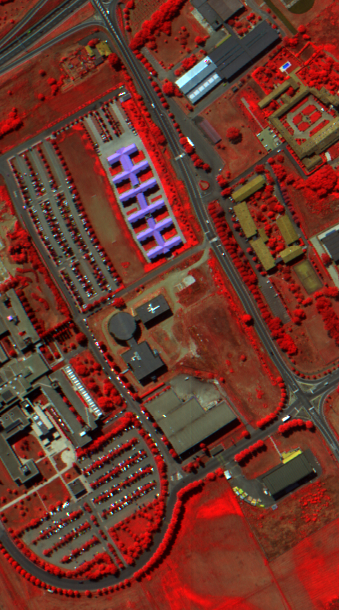}}

    \caption{Classification maps of Pavia University. \textbf{w/} represents multitask learning with spectral knowledge, whereas \textbf{w/o} represents multitask learning without spectral knowledge. PC, IN and SA indicate Pavia Center, Indian Pines and Salinas Valley, respectively. ALL indicates multitask learning with all the data. \cb{The OAs from left to right for (a-f) are 75.25\%, 81.03\%, 79.83\%, 80.42\%, 81.45\% and 80.54\%.}}
    \label{fig:maps}
\end{figure*}

For illustrate purposes, we present classification maps using 10 samples per class with different methods in Fig. \ref{fig:maps}. Compared with \cb{single-task learning}, classification maps generated by multitask learning are more accurate and less salt-and-pepper, especially the one cotrained with all the data.

\subsection{Experiments on the AVIRIS Data Sets}
In the second group of experiments, we conducted multitask learning on the AVIRIS data sets. 
To reduce the atmospheric effects, only the last 160 spectral were applied in the experiments.
When the ROSIS data sets were involved, we repeated the last 57 spectral bands for Pavia University and the last 58 spectral bands for Pavia Center, respectively. In this case, the spectral information was mismatched among data.  
The results of Indian Pines are represented in Table \ref{tab:oa_in}. \cb{Both HResNet and CCNN benefited from multitask learning. For instance, using HResNet, }
the OAs were improved from 39.17\% to 42.44\% with 5 samples per class and 70.76\% to 73.66\% with 30 samples per class. 
Results cotrained with Salinas Valley were the best in all settings, probably because the data were from the same sensor and the spectral knowledge was meaningful across domains. 

The results of Salinas Valley were represented in Table \ref{tab:oa_sa}. Only on this data set that we observe a decrease in terms of OA with multitask learning.  
It might be that this data set has the most complex classification system with 16 classes and a relatively high classification performance. Spectral knowledge from other data sets was unhelpful to this task.

\begin{table*}[!t]
  \centering
  \caption{Overall accuracy along with standard deviation of Indian Pines.}
   \scalebox{0.78}[0.78]{
    \begin{tabular}{c|cc|cccc|cccc|cc}\hline
    \toprule
    NoS/class  & \multicolumn{2}{c|}{\cb{Single-task}} & \multicolumn{4}{c|}{MTL w/ spectral knowledge, \cb{HResNet}} & \multicolumn{4}{c|}{MTL w/o spectral knowledge, \cb{HResNet}} & \multicolumn{2}{c}{\cb{CCNN}} \\
    per domain & \cb{SVM} & \cb{HResNet}   & SA    & PC    & PU    & ALL   & SA & PC & PU & \cb{Inverse-SA} & \cb{Single-task} & \cb{MTL-SA} \\
    \midrule
    5   & \cb{38.29$\pm$4.44} & 39.17$\pm$3.06 & 42.44$\pm$3.24 & 41.09$\pm$3.77 & 40.26$\pm$2.95 & 41.54$\pm$3.20 & 36.17$\pm$5.08 & 40.31$\pm$3.49 & 39.59$\pm$3.10 & \cb{41.09$\pm$2.57} & \cb{42.00$\pm$5.41} & \cb{42.59$\pm$3.85} \\
    10  & \cb{47.05$\pm$3.49} & 48.09$\pm$3.32   & 53.61$\pm$3.97 & 53.73$\pm$3.23 & 52.28$\pm$3.46 & 53.27$\pm$3.31 & 54.62$\pm$3.25 & 52.61$\pm$4.88 & 49.21$\pm$4.56 & \cb{53.78$\pm$3.91} & \cb{54.54$\pm$4.65} & \cb{58.86$\pm$3.16} \\
    15  & \cb{51.05$\pm$2.29} & 56.95$\pm$3.42   & 61.41$\pm$3.42 & 60.99$\pm$2.67 & 59.67$\pm$2.83 & 59.62$\pm$1.28 & 62.18$\pm$1.29 & 60.22$\pm$3.66 & 59.75$\pm$2.95 & \cb{61.99$\pm$2.85} & \cb{61.09$\pm$2.41} & \cb{63.99$\pm$3.93} \\
    20  & \cb{55.59$\pm$1.37} & 64.54$\pm$1.98  & 67.73$\pm$1.86 & 66.25$\pm$2.20 & 64.46$\pm$1.99 & 66.11$\pm$2.11 & 66.53$\pm$1.69 & 60.29$\pm$3.47 & 66.13$\pm$2.14 & \cb{67.27$\pm$1.76}  & \cb{64.91$\pm$2.96} & \cb{69.46$\pm$2.90}\\
    25  & \cb{57.25$\pm$1.61} & 68.17$\pm$1.32 & 70.83$\pm$1.93 & 69.97$\pm$2.10 & 68.04$\pm$1.62 & 70.57$\pm$2.02 & 71.33$\pm$2.72 & 69.76$\pm$2.06 & 69.50$\pm$1.70 & \cb{70.75$\pm$1.38}  & \cb{67.01$\pm$2.12} & \cb{72.81$\pm$2.52} \\
    30  & \cb{59.35$\pm$1.66} & 70.76$\pm$2.07  & 73.66$\pm$1.11 & 72.50$\pm$0.86 & 71.35$\pm$1.26 & 72.66$\pm$1.42 & 72.69$\pm$1.75 & 71.80$\pm$1.59 & 72.60$\pm$1.17 & \cb{73.37$\pm$1.32} & \cb{70.80$\pm$1.78} & \cb{74.82$\pm$1.78} \\
    \bottomrule \hline
    \end{tabular}}%
  \label{tab:oa_in}%
\end{table*}%

\begin{table*}[!t]
  \centering
  \caption{Overall accuracy along with standard deviation of Salinas Valley.}
   \scalebox{0.78}[0.78]{
    \begin{tabular}{c|cc|cccc|cccc|cc} \hline
    \toprule
    NoS/class  & \multicolumn{2}{c|}{\cb{Single-task}} & \multicolumn{4}{c|}{MTL w/ spectral knowledge, \cb{HResNet}} & \multicolumn{4}{c|}{MTL w/o spectral knowledge, \cb{HResNet}} & \multicolumn{2}{c}{\cb{CCNN}} \\
    per domain & \cb{SVM} & \cb{HResNet}    & IN    & PC    & PU    & ALL   & IN & PC & PU & \cb{Inverse-IN} & \cb{Single-task} & \cb{MTL-IN} \\
    \midrule
    5   & \cb{76.21$\pm$3.20} & 80.29$\pm$2.76  & 71.15$\pm$2.75 & 76.75$\pm$2.66 & 72.30$\pm$4.02 & 70.86$\pm$2.78 & 77.63$\pm$3.66 & 79.60$\pm$2.04 & 77.57$\pm$3.87 & \cb{68.75$\pm$2.76}   & \cb{75.36$\pm$3.89} &  \cb{72.85$\pm$4.49}  \\
    10  & \cb{80.63$\pm$3.00} & 81.29$\pm$1.98   & 77.78$\pm$1.55 & 79.58$\pm$1.71 & 76.60$\pm$2.32 & 76.33$\pm$1.09 & 80.03$\pm$1.95 & 80.57$\pm$1.25 & 79.72$\pm$1.37 & \cb{76.99$\pm$1.61} & \cb{80.73$\pm$3.53} &  \cb{82.54$\pm$3.48} \\
    15  & \cb{83.56$\pm$1.79} & 82.84$\pm$2.38   & 80.77$\pm$1.22 & 82.24$\pm$1.86 & 80.15$\pm$2.07 & 79.96$\pm$1.11 & 81.54$\pm$2.81 & 82.34$\pm$2.37 & 81.82$\pm$2.05 & \cb{80.58$\pm$1.61} & \cb{83.72$\pm$3.54} & \cb{83.88$\pm$2.53} \\
    20  & \cb{84.84$\pm$1.27} & 83.78$\pm$1.90  & 82.27$\pm$0.93 & 82.95$\pm$1.41 & 82.29$\pm$1.57 & 82.29$\pm$1.34 & 84.16$\pm$2.14 & 84.09$\pm$1.48 & 83.32$\pm$1.41 &  \cb{82.29$\pm$1.06}   & \cb{84.74$\pm$3.75} & \cb{86.31$\pm$1.97} \\
    25  & \cb{85.35$\pm$1.99} & 85.11$\pm$2.03 & 83.27$\pm$1.39 & 84.23$\pm$1.65 & 83.27$\pm$1.69 & 83.74$\pm$1.25 & 84.29$\pm$1.61 & 84.37$\pm$1.59 & 83.91$\pm$1.56 & \cb{83.94$\pm$1.03} & \cb{87.66$\pm$1.46} &  \cb{87.61$\pm$1.53}\\
    30  & \cb{87.33$\pm$1.65} & 85.63$\pm$1.80  & 84.70$\pm$1.43 & 85.29$\pm$1.18 & 84.02$\pm$1.15 & 84.27$\pm$1.34 & 85.21$\pm$1.46 & 85.96$\pm$1.97 & 84.86$\pm$1.80 & \cb{85.37$\pm$1.39}  &  \cb{87.78$\pm$1.32} &  \cb{87.70$\pm$2.03}\\
    \bottomrule \hline
    \end{tabular}}%
  \label{tab:oa_sa}%
\end{table*}%

\subsection{Computation Time}
Finally, \cb{we compare the computation time} between the proposed method and standard CNN in Table \ref{tab:computation_time}. The experiments were conducted on GTX 1050 6G with 10 samples per class. \cb{Higher accuracy and less computation time are observed in multitask learning (two tasks) compared with two single tasks.}

\begin{table}[htbp]
  \centering
  \caption{Comparison of computation time.}
   \scalebox{0.98}[0.98]{
    \begin{tabular}{cccccc} \hline
    \toprule
          & \multicolumn{3}{c}{\cb{Single-task}} & \multicolumn{2}{c}{Multitask learning} \\
          & OA (\%) & \multicolumn{2}{c}{Time (s)} & OA (\%) & Time (s) \\
    \midrule
    Pavia University & 72.5  & 15.9 & \multirow{2}[1]{*}{\cb{34.6}} & 75.1  & \multirow{2}[1]{*}{\textbf{23.2}} \\
    Pavia Center & 96.6  & 18.7 & & 97.3    &  \\
    Indian Pines & 48.1  & 25.3 & \multirow{2}[1]{*}{\cb{49.7}}  & 53.6  & \multirow{2}[1]{*}{\textbf{32.5}} \\
    Salinas Valley & 81.3  & 24.4 &  & 77.8  &  \\
    \bottomrule \hline
    \end{tabular}}%
  \label{tab:computation_time}%
\end{table}%

\section{Conclusion}
This letter proposes multitask deep learning for hyperspectral image classification, which can utilize samples from multiple data sets and alleviate the overfitting problem from the dilemma between big networks and small samples. 
\cb{The obtained results demonstrate that the proposed method successfully enhances the performance of deep CNNs, especially when training samples are limited. Additionally, the computation time is saved since one multitask network is obtained to classify two or even more hyperspectral data sets in a single training.}
We plan to integrate this method (which belongs to transfer learning) with active learning and semi-supervised learning in the future.


\tiny{
\ifCLASSOPTIONcaptionsoff
  \newpage	
\fi

\bibliographystyle{IEEEtran}
\bibliography{strings}
}




\end{document}